\documentclass[letterpaper]{sig-alternate-2013}

\usepackage{graphicx}
\usepackage{epsfig}
\usepackage[caption=false]{subfig}
\usepackage{placeins}
\usepackage{capt-of}
\usepackage{multirow,makecell}
\usepackage{amsmath}
\usepackage{mathtools}
\usepackage{Definitions}
\usepackage{siunitx}
\usepackage{fixltx2e}
\usepackage{xspace}
\usepackage{color}
\usepackage{pifont}
\usepackage[boxed,figure]{algorithm2e}
\usepackage[table]{xcolor}
\usepackage{array}
\usepackage{enumerate}
\usepackage{paralist}
\usepackage{hhline}

\SetKwInput{KwInput}{Input}
\SetKwInput{KwOutput}{Output}
\newcommand{\cmark}{\ding{51}}
\newcommand{\xmark}{\ding{55}}

\newcommand{\xhdr}[1]{\vspace{0.5mm}\noindent{{\bf #1.}}}

\newcommand{\disc}{disparate mistreatment}
\newcommand{\Disc}{Disparate mistreatment}

\permission{\copyright 2017 International World Wide Web Conference Committee (IW3C2),
published under Creative Commons CC BY 4.0 License.}
\conferenceinfo{WWW 2017,}{April 3--7, 2017, Perth, Australia.}
\copyrightetc{ACM \the\acmcopyr}
\crdata{978-1-4503-4913-0/17/04. \\
http://dx.doi.org/10.1145/3038912.3052660 \\
\includegraphics{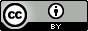}
}

\begin{document}

\title{
Fairness Beyond Disparate Treatment \& Disparate Impact: Learning Classification without Disparate Mistreatment
\titlenote{
 \scriptsize{An open-source code implementation of our scheme is available at: }
\textbf{\url{http://fate-computing.mpi-sws.org/}}
}
}

\numberofauthors{1}
\author{
    \alignauthor Muhammad Bilal Zafar, Isabel Valera, Manuel Gomez Rodriguez, Krishna P. Gummadi\\
    \vspace{1mm}
    \affaddr{Max Planck Institute for Software Systems (MPI-SWS)}
    \vspace{1mm}
    \email{  \{mzafar, ivalera, manuelgr, gummadi\}@mpi-sws.org}
}

\maketitle

\begin{abstract}

Automated data-driven decision making systems are increasingly being
used to assist, or even replace humans in many settings. These systems
function by learning from historical decisions, often taken by
humans. In order to maximize the utility of these systems (or,
classifiers), their training involves minimizing the errors (or,
misclassifications) over the given historical data. However, it is
quite possible that the optimally trained classifier makes decisions
for people belonging to different social groups with different {\it
  misclassification} rates (e.g., misclassification rates for females
are higher than for males), thereby placing these groups at an unfair
disadvantage. To account for and avoid such unfairness, in this paper,
we introduce a new notion of unfairness, {\it disparate mistreatment},
which is defined in terms of misclassification rates. We then propose
intuitive measures of disparate mistreatment for decision
boundary-based classifiers, which can be easily incorporated into
their formulation as convex-concave constraints. Experiments on
synthetic as well as real world datasets show that our methodology is
effective at avoiding disparate mistreatment, often at a small cost in
terms of accuracy.

\end{abstract}

\section{Introduction}\label{sec:intro}
\noindent

The emergence and widespread usage of automated data-driven decision
making systems in a wide variety of applications, ranging from content
recommendations to pretrial risk assessment, has raised concerns about
their potential unfairness towards people with certain
traits~\cite{crawford_ai_disc,bigdatawhitehouse2016,bigdatawhitehouse2014,sweeney_queue}.
Anti-discrimination laws in va\-rious countries prohibit unfair
treatment of in\-di\-vi\-duals based on specific traits, also called
\textit{sensitive} attributes (\eg, gender, race).  These laws
typically distinguish between two different notions of
unfairness~\cite{barocas_2016} namely, \textbf{\textit{disparate
    treatment}}
and \textbf{\textit{disparate impact}}.
More spe\-ci\-fi\-cally, there is disparate treatment when the
decisions an individual user receives change with changes to her
sensitive attribute information, and there is disparate impact when
the decision outcomes disproportionately benefit or hurt members of
certain sensitive attribute value groups.
A number of recent studies~\cite{feldman_kdd15, salvatore_knn,
  icml2013_zemel13}, including our own prior
work~\cite{zafar_fairness}, have focused on de\-signing decision
making systems that avoid one or both of these types of unfairness.

These prior designs have attempted to tackle unfairness in decision
making scenarios where the historical decisions
in the training data are \emph{biased} (\ie, groups of people with
certain sensitive attributes may have historically received unfair treatment)
and there is no ground truth about the \emph{correctness}
of the historical decisions (\ie, one cannot tell whether a historical decision used during the
training phase was right or wrong).
However, when the ground truth
for historical decisions is available,
disproportionately beneficial outcomes for certain sensitive attribute
value groups can be justified and explained by means of the
ground truth. Therefore, disparate impact would not be a suitable notion of
unfairness in such scenarios.

In this paper, we propose an alternative notion of
unfairness, \textbf{\textit{\disc{}}}, especially well-suited for
scenarios where ground truth is available for historical decisions used during
the training phase.
We call a decision ma\-king process to be suffering from \disc{} with
respect to a given sensitive attribute (\eg, race) if the
\emph{misclassification rates differ} for groups of people having different values
of that sensitive attribute (\eg, blacks and whites).
 For example, in the case of the NYPD Stop-question-and-frisk program
(SQF)~\cite{sqf_wiki}, where pedestrians are stopped on the suspicion of possessing an illegal
weapon~\cite{goel_frisk}, having different
weapon dis\-co\-ve\-ry rates
for different races would constitute a case of \disc{}.

\begin{figure*}[ht]
\centering
\resizebox{\textwidth}{!}{

\begin{tabular}{|c|c|c|c|c|c|c|c|c|c|c|c|c|c|}

\cline{1-3} \cline{5-5} \cline{7-9} \cline{11-14}
\multicolumn{3}{ |c| }{\textbf{\small User Attributes}} && {\textbf{\small Ground Truth}} && \multicolumn{3}{ c| }{\textbf{\small Classifier's}} && & \textbf{\small Disp.} & \textbf{\small Disp.} &  \textbf{\small Disp.} \\
\cline{1-3}

\textbf{\small Sensitive} & \multicolumn{2}{ c| }{\textbf{\small Non-sensitive}} && \textbf{\small (Has Weapon)} && \multicolumn{3}{ c| }{\textbf{\small Decision to Stop}} && & \textbf{\small Treat.} & \textbf{\small Imp.} &  \textbf{\small Mist.} \\

\cline{1-3} \cline{7-9}
\textbf{\small Gender}  &  \textbf{\small Clothing Bulge} & \textbf{\small Prox. Crime} && \textbf{\small} && \small $\mathbf{C_1}$ & \small $\mathbf{C_2}$ & \small $\mathbf{C_3}$ && & \textbf{} & \textbf{} &  \textbf{} \\
\cline{1-3} \cline{5-5} \cline{7-9}  \cline{11-14}

Male 1   & 1 & 1  &&  \cmark   &&  1 & 1 & 1  && \multirow{2}{*}{$\mathbf{C_1}$} & \multirow{2}{*}{\xmark} & \multirow{2}{*}{\cmark} & \multirow{2}{*}{\cmark}
\\ \cline{1-3} \cline{5-5} \cline{7-9}

Male 2   & 1 & 0  &&  \cmark   &&  1 & 1 & 0 &&&&&
\\ \cline{1-3} \cline{5-5} \cline{7-9}  \cline{11-14}

Male 3   & 0 & 1  &&  \xmark   &&  1 & 0 & 1  && \multirow{2}{*}{$\mathbf{C_2}$} & \multirow{2}{*}{\cmark} & \multirow{2}{*}{\xmark} & \multirow{2}{*}{\cmark}
\\ \cline{1-3} \cline{5-5} \cline{7-9}

Female 1 & 1 & 1  &&  \cmark   &&  1 & 0 & 1 &&&&&
\\ \cline{1-3} \cline{5-5} \cline{7-9}  \cline{11-14}

Female 2 & 1 & 0  &&  \xmark   &&  1 & 1 & 1  && \multirow{2}{*}{$\mathbf{C_3}$} & \multirow{2}{*}{\cmark} & \multirow{2}{*}{\xmark} & \multirow{2}{*}{\xmark}
\\ \cline{1-3} \cline{5-5} \cline{7-9}

Female 3 & 0 & 0  &&  \cmark   &&  0 & 1 & 0 &&&&&
\\ \cline{1-3} \cline{5-5} \cline{7-9}  \cline{11-14}

\end{tabular}
}

\caption{
\small{
Decisions of three fictitious classifiers ($\mathbf{C_1}$, $\mathbf{C_2}$ and $\mathbf{C_3}$) on whether ($1$) or not ($0$) to stop a pedestrian on the suspicion of possessing an illegal weapon. Gender is a sensitive attribute, whereas the other two attributes (suspicious bulge in clothing and proximity to a crime scene) are non-sensitive. Ground truth on whether the person is actually in possession of an illegal weapon is also shown.
}
}
\vspace{-3ex}
\label{table:examples-disc}
\end{figure*}

In addition to \textit{all} misclassifications in general, depending
on the application scenario, one might want to measure \disc{} with
respect to different kinds of misclassifications.
For example, in pretrial risk assessments,
the decision making process might only be required to ensure that
the false positive rates are equal for all groups, since it may be more acceptable to let
a guilty person go, rather than incarcerate an innocent
person.
\footnote{
\scriptsize{\textit{``It is better that ten guilty persons escape than that one innocent suffer''}---William Blackstone}
}
On the other hand, in loan approval systems, one might instead favor a
decision making process in which the false negative rates are equal,
to ensure that deserving (positive class) people with a certain
sensitive attribute value are not denied (negative class) loans
disproportionately.
Similarly, depending on the application scenario at hand, and the cost of the type of misclassification, one may choose to measure  \disc{} using
false discovery and false omission rates, instead of false positive and false negative rates (see Table~\ref{table:error-rates}).

In the remainder of the paper, we first formalize disparate treatment, disparate impact and disparate mistreatment in the context of (binary) classification.
Then, we introduce intuitive measures of \disc{} for
decision boundary-based classifiers and show that, for a wide variety
of linear and nonlinear classifiers, these measures can be incorporated
into their formulation as convex-concave constraints. The resulting
formulation can be solved efficiently using recent advances in
convex-concave programming~\cite{boyd_concave_convex}.
Finally, we experiment with synthetic as well as real world datasets and show that our
methodology can be effectively used to avoid \disc{}.

\section{Background and Related Work}

\noindent In this section, we first elaborate on the three different notions of unfairness
in automated decision making systems using an illustrative example and then provide
an overview of the related literature.

\vspace{1.5mm}
\xhdr{Disparate mistreatment} Intuitively, disparate mistreatment can arise in any automated
decision making system whose outputs (or decisions) are not perfectly (\ie, 100\%) accurate.
For example, consider a decision making system that uses a logistic regression classifier to
provide binary outputs (say, positive and negative) on a set of people.
If the items in the training data with positive and negative class labels
are not linearly separable, as is often the
case in many real-world application scenarios, the system will misclassify (\ie, produce false positives,
false negatives, or both, on) some people.
In this context, the misclassification rates may be different for groups of people having different values
of sensitive attributes (\eg, males and females; blacks and whites) and thus disparate mistreatment
may arise.

Figure~\ref{table:examples-disc} provides an example of decision making systems (classifiers) with and
without disparate mistreatment.
In all cases, the classifiers need to decide whether to stop a pedestrian---on the suspicion
of possessing an illegal weapon---using a set of
features such as bulge in clothing and proximity to a crime scene.
The ``ground truth'' on whether
a pedestrian actually possesses an illegal weapon is also shown.
We show decisions made by three different classifiers $\mathbf{C_1}$, $\mathbf{C_2}$ and $\mathbf{C_3}$.
We deem $\mathbf{C_1}$ and $\mathbf{C_2}$ as unfair due to disparate mistreatment because their rate of
erroneous decisions for males and females are different: $\mathbf{C_1}$ has different false negative rates for
males and females ($0.0$ and $0.5$, respectively), whereas $\mathbf{C_2}$ has different false positive rates
($0.0$ and $1.0$) as well as different false negative rates ($0.0$ and $0.5$) for males and females.

\vspace{1.5mm}
\xhdr{Disparate treatment}
In contrast to disparate mistreatment, disparate treatment arises when a decision making system
 provides different outputs for groups of people with the same
(or similar) values of non-sensitive attributes (or features) but different values of sensitive
attributes.

In Figure~\ref{table:examples-disc}, we deem $\mathbf{C_2}$ and $\mathbf{C_3}$ to be unfair
due to disparate treatment since $\mathbf{C2}$'s ($\mathbf{C3}$'s) decisions for $Male\ 1$ and
$Female\ 1$ ($Male\ 2$ and $Female\ 2$) are different even though they have the same values
of non-sensitive attributes.
Here, disparate treatment corresponds to the very intuitive notion of fairness: two otherwise
similar persons should not be treated differently solely because of a difference in gender.

\vspace{1.5mm}
\xhdr{Disparate impact} Finally, disparate impact arises when a decision
making system provides outputs that benefit (hurt) a group of people sharing a value
of sensitive attribute more frequently than other groups of people.

In Figure~\ref{table:examples-disc}, assuming  that a pedestrian benefits from a
decision of not being stopped, we deem $\mathbf{C_1}$ as unfair due to disparate
impact because the fraction of males and females that were stopped are different ($1.0$
and $0.66$, respectively).

\vspace{1.5mm}
\xhdr{Application scenarios for disparate impact vs. disparate mistreatment} Note that unlike
in the case of disparate mistreatment, the notion of disparate impact is independent of the ``ground truth''
information about the decisions, \ie, whether or not the decisions are correct or valid.
Thus, the notion of disparate impact is particularly appealing in application scenarios where ground
truth information for decisions does not exist and the historical decisions used during training are not
reliable and thus cannot be trusted.
Unreliability of historical decisions for automated decision making systems is particularly concerning in
scenarios like recruiting or loan approvals, where biased judgments by humans in the past may be used
when training classifiers for the future.
In such application scenarios, it is hard to distinguish correct and incorrect decisions, making it hard to assess
or use disparate mistreatment as a notion of fairness.

However, in  scenarios where ground truth information for decisions can be obtained, disparate impact
can be quite misleading as a notion of fairness. That is, in scenarios where the validity of decisions can be reliably
ascertained, it would be possible to distinguish disproportionality in decision outcomes for sensitive groups that
arises from justifiable reasons (\eg, qualification of the candidates) and disproportionality that arises for
non-justifiable reasons (\ie, discrimination against certain groups).
By requiring decision outcomes to be proportional, disparate impact risks introducing reverse-discrimination
against quali\-fied candidates. Such practices have previously been deemed unlawful by courts (\textit{Ricci vs. DeStefano, 2009}).
In contrast, when the correctness of decisions can be determined, disparate mistreatment can not only be
accurately assessed, but also avoids reverse-discrimination, making it a more appealing notion of fairness.

\vspace{1.5mm}
\xhdr{Related Work}
There have been a number of studies, including our own prior work~\cite{zafar_fairness}, proposing methods for
detecting~\cite{feldman_kdd15,salvatore_knn,pedreschi_discrimination,salvatore_survey}
and removing~\cite{Dwork2012,feldman_kdd15,goh_nips2016,kamiran_sampling,kamishima_regularizer,pedreschi_discrimination,zafar_fairness,icml2013_zemel13}
unfairness when it is defined in terms of disparate treatment, disparate impact or both.
However, as pointed out earlier, the disparate impact notion might be less
meaningful in scenarios where ground truth decisions are available.

A number of previous studies have pointed out racial disparities in both automated~\cite{propublica_story}
as well as human~\cite{goel_frisk,fdr_benchmarking} decision making systems related to criminal justice.
For example, a recent work by Goel et al.~\cite{goel_frisk} detects racial disparities in NYPD SQF
program, inspired by a notion of unfairness similar to our notion of disparate
mistreatment.
More specifically, it uses
ground truth (stops leading to successful discovery of an illegal weapon on the
suspect) to show that blacks were treated unfairly since false
positive rates in stops were higher for them than for whites.
The study's findings provide further justification for the need for
data-driven decision making systems without \disc{}.

A recent work by Hardt et al.~\cite{hardt_nips16} (concurrently conducted with our work) proposes a method
to achieve a fairness notion equivalent to our notion of \disc{}.
 This method
works by post-processing the probability estimates of an unfair classifier to learn
 different decision thresholds for different sensitive attribute value groups, and applying these group-specific thresholds at decision making time.
Since this method requires the sensitive attribute information at decision time,
it cannot be used in cases where sensitive attribute information is unavailable (\eg, due to privacy reasons) or prohibited
from being used due to disparate treatment laws~\cite{barocas_2016}.

\vspace{-3mm}
\section{Formalizing Notions of Fairness} \label{sec:format_setting}
\noindent
In a binary classification task, the goal is to learn a mapping $f(\mathbf{x})$ between user feature vectors $\mathbf{x} \in \RR^d$ and class labels $y \in \{-1, 1\}$. Learning this mapping
is often achieved by finding a decision boundary $\thetab^*$ in the feature space that mi\-ni\-mizes a certain loss $L(\thetab)$, \ie, $\thetab^{*} = \argmin_{\thetab} L(\thetab)$,
computed on a training dataset $\mathcal{D} = \{ (\mathbf{x}_i, y_i) \}_{i=1}^{N}$.
Then, for a given \emph{unseen} feature vector $\mathbf{x}$,
the classifier predicts the class label
$\hat{y} = f_{\thetab^{*}}(\mathbf{x}) = 1$ if $d_{\thetab^{*}}(\mathbf{x}) \geq 0$ and
$\hat{y} = -1$
other\-wise, where $d_{\thetab^{*}}(\mathbf{x})$ denotes the signed distance from
$\mathbf{x}$ to the decision boundary.
Assume that each user has an associated sensitive feature $z$. For ease of exposition,
we assume $z$ to be binary, \ie, $z \in \{0, 1\}$.  However, our setup can be easily generalized to categorical
as well as multiple sensitive features.

\begin{table}
\centering
\resizebox{\columnwidth}{!}{
\begin{tabular}{cc|c|c|c}
\cline{3-4}\noalign{\vspace{0.5pt}}
&&  \multicolumn{2}{ c|  }{\cellcolor{gray!25} \textbf{Predicted Label}}
\\ \cline{3-4}\noalign{\vspace{0.5pt}}
&
& \cellcolor{gray!25} $\hat{y} = 1$
& \cellcolor{gray!25} $\hat{y} = -1$
&  \\ \hline
\multicolumn{1}{ |c  }{\cellcolor{gray!25}}
&  \multicolumn{1}{ |c|  }{\raisebox{1em}{\rotatebox[origin=c]{90}{\cellcolor{gray!25} $y = 1$}} }
&  {\raisebox{1em}{\rotatebox[origin=c]{0}{True positive}} }
&  {\raisebox{1em}{\rotatebox[origin=c]{0}{False negative}} }
&  \multicolumn{1}{ c|  }{\rotatebox[origin=c]{0}{\cellcolor{blue!25} \shortstack{$P (\hat{y} \neq y | y = 1)$\\ False \\Negative Rate}}}
\\ \cline{2-5}
\hhline{|>{\arrayrulecolor{gray!25}}->{\arrayrulecolor{black}}---}
\multicolumn{1}{ |c  }{\multirow{-5}{*}{\rotatebox[origin=c]{90}{\cellcolor{gray!25} \textbf{True Label}}}}
& \multicolumn{1}{ |c|  }{\raisebox{1em}{\rotatebox[origin=c]{90}{\cellcolor{gray!25} $y = -1$} }}
&  {\raisebox{1em}{\rotatebox[origin=c]{0}{False positive}} }
&  {\raisebox{1em}{\rotatebox[origin=c]{0}{True negative}} }
&  \multicolumn{1}{ c|  }{\rotatebox[origin=c]{0}{\cellcolor{blue!25} \shortstack{$P (\hat{y} \neq y | y = -1)$\\ False \\Positive Rate}}}
\\ \hline
&
&  \multicolumn{1}{ c|  }{\rotatebox[origin=c]{0}{\cellcolor{cyan!25} \shortstack{$P (\hat{y} \neq y | \hat{y} = 1)$\\ False \\Discovery Rate}}}
&  \multicolumn{1}{ c|  }{\rotatebox[origin=c]{0}{\cellcolor{cyan!25} \shortstack{$P (\hat{y} \neq y | \hat{y} = -1)$\\ False \\Omission Rate}}}
&  \multicolumn{1}{ c|  }{\rotatebox[origin=c]{0}{\cellcolor{orange!25} \shortstack{$P (\hat{y} \neq y)$\\ Overall \\Misclass. Rate}}}
\\ \cline{3-5}
\end{tabular}
}
\caption{
\small{
In addition to the overall misclassification rate, error rates can be measured in two different ways: false negative rate and false positive rate are defined as fractions over the \textit{class distribution in the ground truth labels}, or true labels. On the other hand, false discovery rate and false omission rate are defined as fractions over the \textit{class distribution in the predicted labels}.
}
}
\vspace{-5mm}
\label{table:error-rates}
\end{table}
\setlength\tabcolsep{6pt}

Given the above terminology, we can formally express the absence of disparate treatment, disparate impact and \disc{} as
follows:

\vspace{2mm}
\noindent {\bf Existing notion 1: Avoiding disparate treatment.} A binary classifier does not suffer from disparate treatment if:
\begin{equation} \label{eq:disparate_treatment}
P(\hat{y} | \mathbf{x}, z) = P(\hat{y} | \mathbf{x}),
\end{equation}
\ie, if the probability that the classifier outputs a specific value of $\hat{y}$ given a feature vector $\mathbf{x}$ does not change after observing
the sensitive feature $z$, there is no disparate treatment.

\vspace{2mm}
\noindent {\bf Existing notion 2: Avoiding disparate impact.} A binary classifier does not suffer from disparate impact if:
\begin{equation} \label{eq:disparate_impact}
P(\hat{y} = 1 | z = 0) = P(\hat{y} = 1 | z = 1),
\end{equation}
\ie, if the probability that a classifier assigns a user to the positive class, $\hat{y} = 1$, is the same for both values of the sensitive feature
$z$, then there is no disparate impact.

\vspace{2mm}
\noindent {\bf New notion 3: Avoiding disparate mistreatment.}
A binary classifier does not suffer from \disc{}  if the misclassification rates for different groups of people having different values of the sensitive feature $z$ are the same. Table~\ref{table:error-rates} describes various ways of measuring misclassification rates.
Specifically, misclassification rates can be measured as fractions over the \textit{class distribution in the ground truth labels}, \ie, as false positive and false negative rates, or over the \textit{class distribution in the predicted labels}, \ie, as false omission and false discovery rates.
\footnote{
\scriptsize{
In prediction tasks where a positive prediction entails a large cost (\eg, cost involved in the treatment of a disease)
one might be more interested in measuring error rates as fractions over the class distribution in the \textit{predicted labels}, rather than over the class distribution in the \textit{ground truth labels}, \eg, to ensure that the false discovery rates, instead of false positive rates, for all groups are the same.
}
}
Consequently, the absence of \disc{} in a binary
classification task can be specified with respect to the different misclassification measures as follows:\\

\noindent {\it overall misclassification rate (OMR)}:
\vspace{-2mm}
\begin{flalign}
P(\hat{y} \neq y | z = 0) = P(\hat{y} \neq y | z = 1), \label{eq:prob_misclass}
\end{flalign}
\noindent {\it false positive rate (FPR)}:
\begin{flalign}
P(\hat{y} \neq y | z = 0, y = -1) = P(\hat{y} \neq y | z = 1, y = -1), \vspace{-10mm}\label{eq:prob_misclass_yminus1}
\end{flalign}

\noindent {\it false negative rate (FNR)}:
\begin{flalign}
P(\hat{y} \neq y | z = 0, y = 1) = P(\hat{y} \neq y | z = 1, y = 1), \label{eq:prob_misclass_y1}
\end{flalign}

\noindent {\it false omission rate (FOR)}:
\begin{flalign}
P(\hat{y} \neq y | z = 0, \hat{y} = -1) = P(\hat{y} \neq y | z = 1, \hat{y} = -1),  \vspace{-10mm}\label{eq:prob_misclass_yhatminus1}
\end{flalign}

\noindent {\it false discovery rates (FDR)}:
\begin{flalign}
P(\hat{y} \neq y | z = 0, \hat{y} = 1) = P(\hat{y} \neq y | z = 1, \hat{y} = 1). \label{eq:prob_misclass_yhat1}
\end{flalign}

\noindent
In the following section, we introduce a method to eliminate \disc{} from decision boundary-based classifiers when \disc{} is defined in terms of overall misclassification rate, false positive rate and false negative rate.
Eliminating \disc{} when it is defined in terms of false discovery rate and false omission rate presents significant additional challenges due to computational complexities involved and we leave it as a direction to be thoroughly explored in a future work.

\xhdr{Satisfying multiple fairness notions simultaneously}
In certain application scenarios, it might be desirable to satisfy
more than one notion of fairness defined above in
Eqs.~(\ref{eq:disparate_treatment}-\ref{eq:prob_misclass_yhat1}). In
this paper, we consider scenarios where we attempt to avoid disparate
treatment as well as disparate mistreatment measured as overall misclassification rate, false
positive rate and false negative rate {\it simultaneously}, \ie, satisfy
Eqs.~(\ref{eq:disparate_treatment}, \ref{eq:prob_misclass}-\ref{eq:prob_misclass_y1}).

Some recent works~\cite{dimpact_fpr,kleinberg_itcs16} have
investigated the impossibility of simultaneously satisfying multiple notions of
fairness.
Chouldechova~\cite{dimpact_fpr} and Kleinberg et
al.~\cite{kleinberg_itcs16}, show that, when the fraction of users
with positive class labels differ between members of different
sensitive attribute value groups, it is impossible to construct
classifiers that are equally {\it well-calibrated} (where
well-calibration essentially measures the false discovery and false omission
rates of a classifier) and also satisfy the equal false positive and
false negative rate criterion (except for a ``dumb'' classifier that
assign all examples to a single class).  These results suggest that
satisfying all five criterion of \disc{}
(Table~\ref{table:error-rates}) simultaneously is impossible when the underlying
distribution of data is different for different groups. However,
in practice, it may still be interesting to explore the best, even if
imperfect, extent of fairness a classifier can achieve. In the next
section, we allow for bounded imperfections in our new fairness
notions by allowing the left- and right-sides of
Eqs.~(\ref{eq:prob_misclass}-\ref{eq:prob_misclass_y1}) to differ
by no more than a threshold $\epsilon$.

\vspace{-2mm}
\section{Classifiers without Disparate \\Mistreatment} \label{sec:methodology}
\noindent
In this section, we describe how to train decision boundary-based classifiers (\eg, logistic regression, SVMs) that do not suffer from \disc{}.
These classifiers ge\-ne\-ra\-lly learn the optimal decision boundary by minimizing a convex loss $L(\thetab)$. The convexity of $L(\thetab)$ ensures
that a global optimum can be found \textit{efficiently}.
In order to ensure that the learned boundary is fair---it does not suffer from \disc{}---one could incorporate the appropriate condition from Eqs.~(\ref{eq:prob_misclass}-\ref{eq:prob_misclass_y1}) (based on which kind of misclassifications \disc{} is being defined for) into the classifier formulation. For example:
\begin{equation}\label{eq:cons_misclass_prob}
	\begin{array}{ll}
		\mbox{minimize} & L(\thetab) \\
		\mbox{subject to}
			& P(\hat{y} \neq y | z = 0) - P(\hat{y} \neq y | z = 1) \leq \epsilon, \\
			& P(\hat{y} \neq y | z = 0) - P(\hat{y} \neq y | z = 1) \geq -\epsilon, \\
	\end{array}
\end{equation}
where $\epsilon \in \RR^+$ and the smaller $\epsilon$ is, the more fair the decision boundary would be. The above formulation ensures that the classifier chooses the optimal decision boundary \textit{within} the space of fair boundaries specified by the constraints.
However, since the conditions in Eqs.~(\ref{eq:prob_misclass}-\ref{eq:prob_misclass_y1}) are, in general, non convex,
solving the constrained optimization problem defined by~\eqref{eq:cons_misclass_prob} seems difficult.

To overcome the above difficulty, we propose a tractable proxy, inspired by the disparate impact proxy proposed by
Zafar et al.~\cite{zafar_fairness}. In particular, we propose to measure \disc{} using the covariance between the users'{}
sensitive attributes and the signed distance between the feature vectors of misclassified users and the classifier decision
boun\-dary, \ie:
\begin{eqnarray} \label{eq:fairness-proxy}
\nonumber
\cov(z, g_{ \thetab}(y, \mathbf{x})) &=&   \EE[ (z - \bar{z}) (g_{ \thetab}(y, \mathbf{x}) -  \bar{g}_{\thetab}(y, \mathbf{x})) ] \\
& \approx& \frac{1}{N} \sum_{(\mathbf{x},y,z) \in \mathcal{D}} \left(z - \bar{z}\right) g_{\thetab}(y, \mathbf{x}) ,
\end{eqnarray}
\noindent
where the term \ $\mathbb{E}[ (z - \bar{z}) ]  \bar{g}_{\thetab}(\mathbf{x})$ cancels out since $\mathbb{E}[ (z - \bar{z}) ] = 0$ and
the function $g_{\thetab}(y,\mathbf{x})$ is defined as:
\begin{align}
g_{ \thetab}(y, \mathbf{x}) &= min(0, y d_{\thetab}(\mathbf{x})), \label{eq:g_misclass} \\
g_{\thetab}(y,\mathbf{x}) &= min\left(0, \frac{1-y}{2} y d_{\thetab}(\mathbf{x})\right),  \mbox{ or } \label{eq:g_fpr} \\
g_{\thetab}(y, \mathbf{x}) &= min\left(0, \frac{1+y}{2} y d_{\thetab}(\mathbf{x})\right), \label{eq:g_fnr}
\end{align}
which approximates, respectively, the conditions in Eqs.~(\ref{eq:prob_misclass}-\ref{eq:prob_misclass_y1}).
Note that, if a decision boundary satisfies Eqs.~(\ref{eq:prob_misclass}-\ref{eq:prob_misclass_y1}), the covariance defined above for that
boundary will be close to zero, \ie, $\cov(z, g_{\thetab}(y, \mathbf{x}))\approx 0$.
Moreover, in linear models for classification, such as logistic regression or linear SVMs, the decision boun\-dary is simply the
hyperplane defined by $\thetab^{T} \mathbf{x} = 0$, therefore, $d_{\thetab}(\mathbf{x}) = \thetab^{T} \mathbf{x}$.

Given the above proxy, one can rewrite~\eqref{eq:cons_misclass_prob} as:
\begin{equation}\label{eq:cons_misclass_non_convex}
	\begin{array}{ll}
		\mbox{minimize} & L(\thetab) \\
		\mbox{subject to}
			& \frac{1}{N} \sum_{(\mathbf{x},y,z) \in \mathcal{D}} \left(z - \bar{z}\right)  g_{\thetab}(y, \mathbf{x})  \leq c, \\
			& \frac{1}{N} \sum_{(\mathbf{x},y,z) \in \mathcal{D}} \left(z - \bar{z}\right)  g_{\thetab}(y, \mathbf{x})  \geq -c,
	\end{array}
\end{equation}
where the covariance threshold $c \in \RR^+$ controls how \emph{adherent} to \disc{} the boundary should be.
\footnote{
\scriptsize{Note that if one wants to have \emph{both} equal false positive and equal false negative rates, one can apply separate constraints with $g_{\thetab}(y,\mathbf{x})$ defined in both~\eqref{eq:g_fpr} and~\eqref{eq:g_fnr}.}
}

\vspace{1.5mm}
\xhdr{Solving the problem efficiently}
While the constraints proposed in~\eqref{eq:cons_misclass_non_convex} can be an effective proxy for fairness, they are still
non-convex, making it challenging to efficiently solve the optimization problem in~\eqref{eq:cons_misclass_non_convex}.
Next, we will convert these constraints into a Disciplined
Convex-Concave Program (DCCP), which can be solved efficiently by
leveraging recent advances in convex-concave programming~\cite{boyd_concave_convex}.

First, consider the constraint described in~\eqref{eq:cons_misclass_non_convex}, \ie,
\begin{align} \label{eq:cons_nonconvex}
\sum_{ (\mathbf{x},y,z) \in \mathcal{D}} \left(z - \bar{z}\right)  g_{\thetab}(y, \mathbf{x})  \sim c,
\end{align}
where $\sim$ may denote `$\geq$' or `$\leq$'. Also, we drop the constant number $\frac{1}{N}$ for the sake of simplicity.
Since the sensitive feature $z$ is binary, \ie, $z\in \{0,1\}$, we can split the sum in the above expression into two terms:
\begin{align} \label{eq:cons_nonconvex_1}
\sum_{ (\mathbf{x},y) \in \mathcal{D}_0 } \left(0 - \bar{z}\right)  g_{\thetab}(y, \mathbf{x}) + \sum_{ (\mathbf{x},y) \in \mathcal{D}_1 } \left(1 - \bar{z}\right)  g_{\thetab}(y, \mathbf{x})  \sim c,
\end{align}
where $\mathcal{D}_0$ and $\mathcal{D}_1$ are the subsets of the training dataset $\mathcal{D}$ taking values $z=0$ and $z=1$, respectively.
Define $N_0 = |\mathcal{D}_0|$ and $N_1 = |\mathcal{D}_1|$, then one can write $\bar{z} = \frac{(0 \times N_0) + (1 \times N_1)}{N} = \frac{N_1}{N}$ and rewrite~\eqref{eq:cons_nonconvex_1} as:
\begin{align} \label{eq:cons_nonconvex_2}
\frac{-N_1}{N} \sum_{ (\mathbf{x},y) \in \mathcal{D}_0 }  g_{\thetab}(y, \mathbf{x}) + \frac{N_0}{N} \sum_{ (\mathbf{x},y) \in \mathcal{D}_1 }  g_{\thetab}(y, \mathbf{x})  \sim c,
\end{align}
which, given that  $g_{ \thetab}(y, \mathbf{x})$ is convex in $\thetab$, results into a convex-concave (or, difference of convex) function.

Finally, we can rewrite the problem defined by~\eqref{eq:cons_misclass_non_convex} as:
\begin{equation}\label{eq:cons_misclass_convex}
	\begin{array}{ll}
		\mbox{minimize} & L(\thetab) \\
		\mbox{subject to}
			& \frac{-N_1}{N} \sum_{ (\mathbf{x},y) \in \mathcal{D}_0 }  g_{\thetab}(y, \mathbf{x}) \\ & + \frac{N_0}{N} \sum_{ (\mathbf{x},y) \in \mathcal{D}_1 }  g_{\thetab}(y, \mathbf{x})  \leq c \\
			& \frac{-N_1}{N} \sum_{ (\mathbf{x},y) \in \mathcal{D}_0 }  g_{\thetab}(y, \mathbf{x}) \\ & + \frac{N_0}{N} \sum_{ (\mathbf{x},y) \in \mathcal{D}_1 }  g_{\thetab}(y, \mathbf{x})  \geq -c,
	\end{array}
\end{equation}
which is a Disciplined Convex-Concave Program (DCCP) for any convex loss $L(\thetab)$, and can be efficiently solved using well-known
heuristics such as the one proposed by Shen et al.~\cite{boyd_concave_convex}.
Next, we particularize the formulation given by~\eqref{eq:cons_misclass_convex} for a logistic regression
classifier~\cite{bishop2006pattern}.
\footnote{
\scriptsize{
Our fairness constraints can be easily incorporated to other boundary-based classifiers such as (non)linear SVMs.
}
}

\vspace{1.5mm}
\xhdr{Logistic regression without disparate mistreatment}
In logistic regression, the optimal decision boundary $\thetab^{*}$ can be found by solving a maximum likelihood problem of the form
$\thetab^{*} = \argmin_{\thetab}  - \sum_{i =1}^N \log p(y_i | \mathbf{x}_i, \thetab)$ in the training phase.
Hence,  a fair logistic regressor can be trained by solving the following constrained optimization problem:
\begin{equation}\label{eq:cons_misclass_convex_logreg}
	\begin{array}{ll}
		\mbox{minimize} & - \sum_{ (\mathbf{x},y) \in \mathcal{D} } \log p(y_i | \mathbf{x}_i, \thetab) \\
		\mbox{subject to}
			& \frac{-N_1}{N} \sum_{ (\mathbf{x},y) \in \mathcal{D}_0 }  g_{\thetab}(y, \mathbf{x}) \\ & + \frac{N_0}{N} \sum_{ (\mathbf{x},y) \in \mathcal{D}_1 }  g_{\thetab}(y, \mathbf{x})  \leq c \\
			& \frac{-N_1}{N} \sum_{ (\mathbf{x},y) \in \mathcal{D}_0 }  g_{\thetab}(y, \mathbf{x}) \\ & + \frac{N_0}{N} \sum_{ (\mathbf{x},y) \in \mathcal{D}_1 }  g_{\thetab}(y, \mathbf{x})  \geq -c.
	\end{array}
\end{equation}

\vspace{1.5mm}
\xhdr{Simultaneously removing disparate treatment}
Note that the above formulation for removing \disc{} provides the flexibility to remove disparate treatment as
well. That is,
since our formulation does not require the sensitive attribute information at decision time, by keeping the features $\mathbf{x}$
disjoint from sensitive attribute $z$, one can remove \disc{} and disparate treatment simultaneously.

\vspace{-2mm}
\section{Evaluation}\label{sec:eval}

\noindent
In this section, we conduct experiments on synthetic as well as real world datasets to evaluate the effectiveness of our scheme in controlling \disc{}. To this end, we first generate several \textit{synthetic} datasets that illustrate diffe\-rent variations of \disc{} and
show that our method can effectively remove \disc{} in each of the variations, often at a small cost on accuracy. Then, we conduct experiments on the ProPublica COMPAS dataset \cite{propublica_compas} to show the effectiveness of our method on a \textit{real world} dataset. In both the synthetic and real-world datasets, we compare the performance of our scheme with a baseline algorithm
and a recently proposed method~\cite{hardt_nips16}.

All of our experiments are conducted using logistic regression classifiers.
To ensure the robustness of the experimental findings, for all of the datasets, we repeatedly (five times) split the
data uniformly at random into train (50\%) and test (50\%) sets and report the average statistics for accuracy and fairness.

\xhdr{Evaluation metrics}
In this evaluation, we consider that one wants to remove \disc{} when it is measured in terms of false positive rate and false negative rate (Eqs.~\eqref{eq:prob_misclass_yminus1} and~\eqref{eq:prob_misclass_y1}).
Specifically, We quantify the \disc{} incurred by a classifier as:
\begin{align}
D_{FPR} &=  P(\hat{y} \neq y | z = 0, y = -1) - P(\hat{y} \neq y | z = 1, y = -1), \nonumber \\
D_{FNR}  &=  P(\hat{y} \neq y | z = 0, y = 1) - P(\hat{y} \neq y | z = 1, y = 1),   \nonumber
\end{align}
where the closer the values of $D_{FPR}$ and $D_{FNR}$ to $0$, the lower the degree of \disc{}.

\begin{figure*}[ht]
 \centering
    \subfloat[Cov. vs FPR]{\includegraphics[trim={7.5cm 0 0 0}, height=0.6\columnwidth, angle=-90]{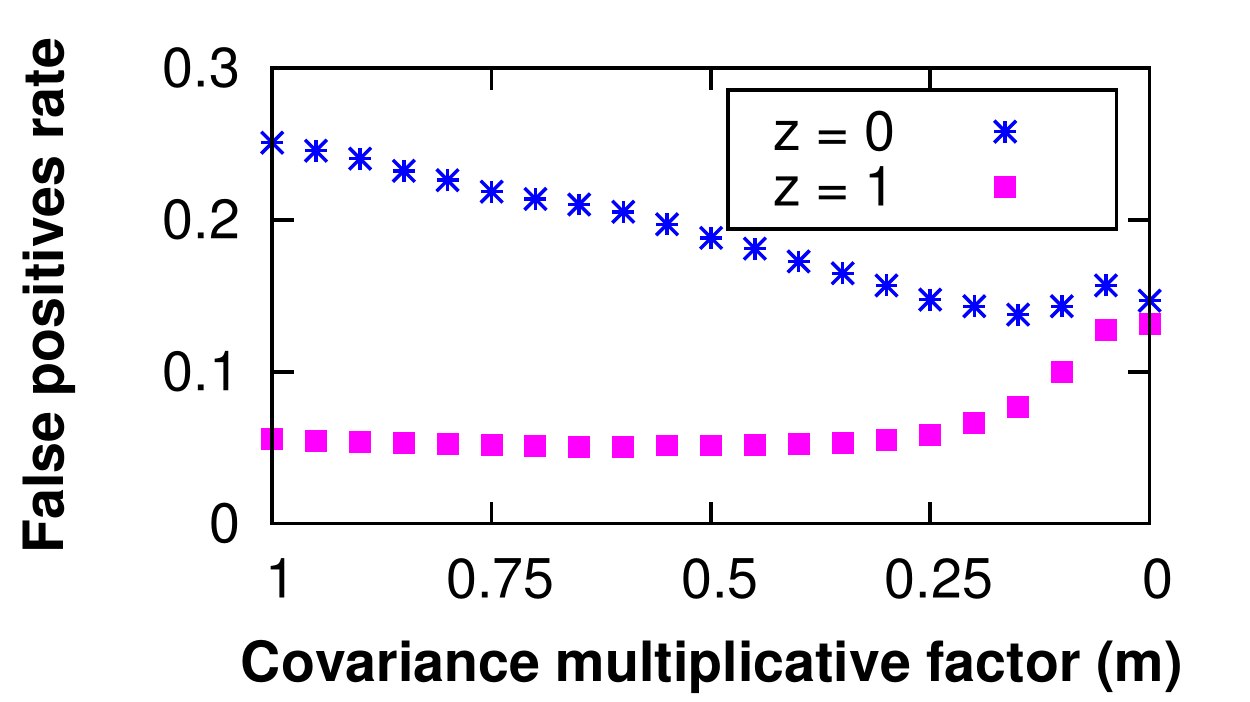}}
    \subfloat[Fairness vs. Acc.]{\includegraphics[trim={7.5cm 0 0 0}, height=0.6\columnwidth, angle=-90]{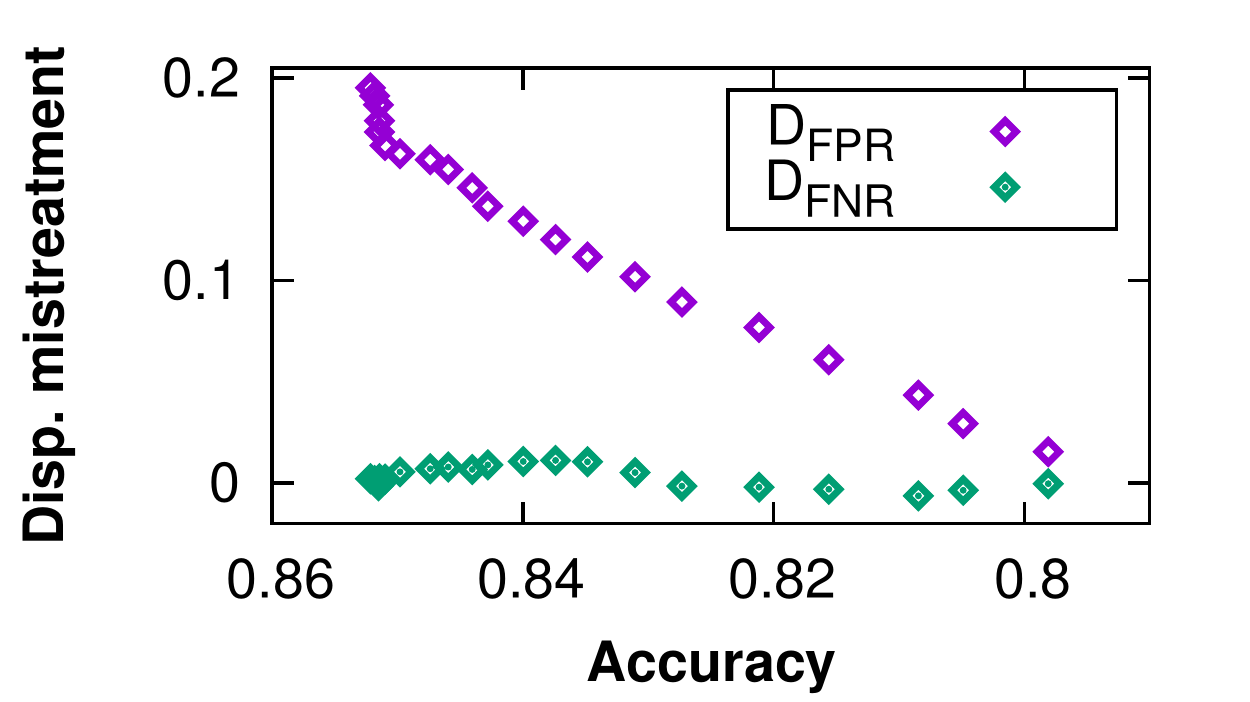}}
    \hspace{10ex}
    \subfloat[Boundaries]{\includegraphics[trim={5cm 0 0 4cm}, width=0.35\columnwidth]{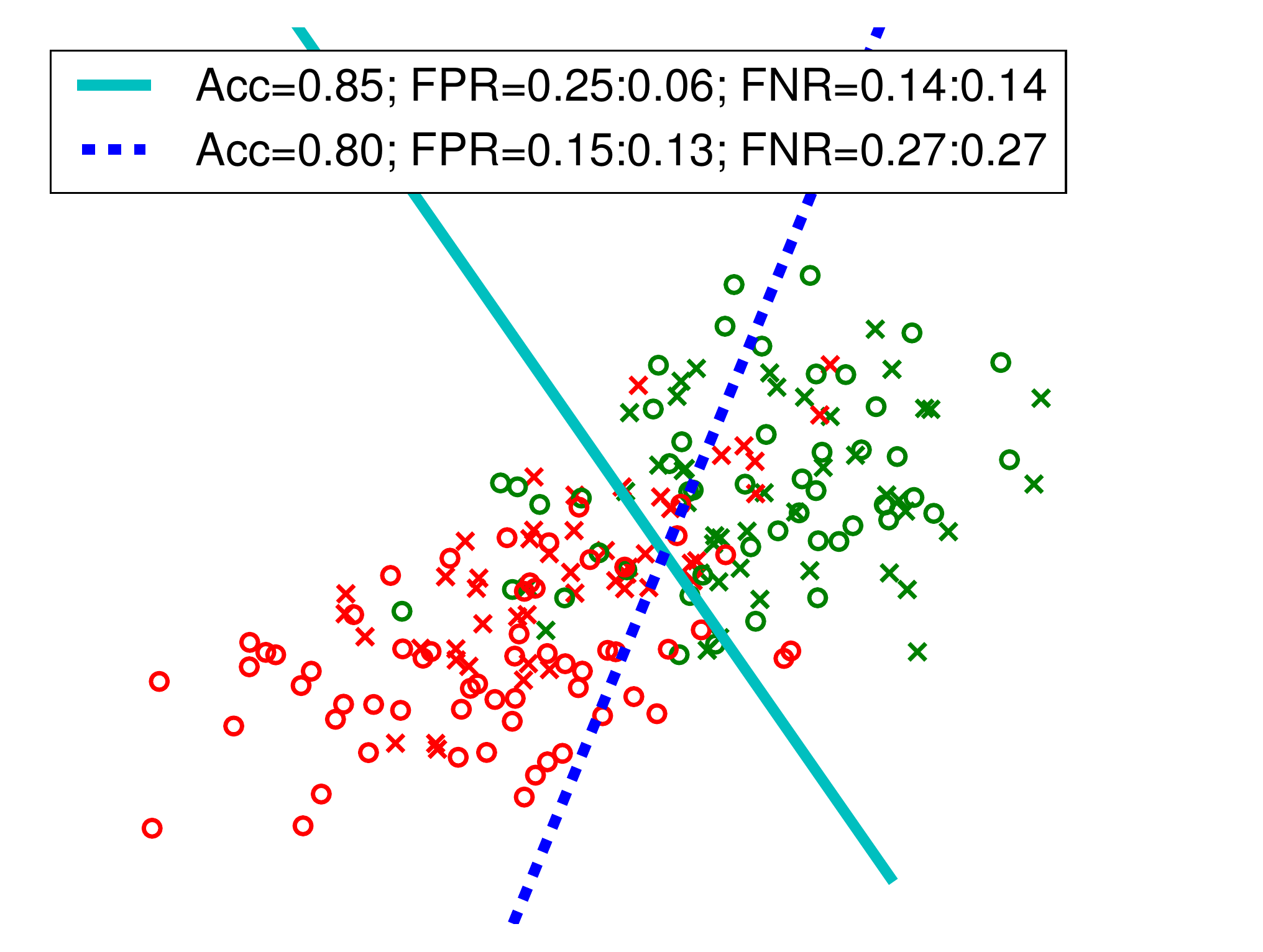}}
    \caption{\small{[Synthetic data] Panel (a) shows that decreasing the covariance threshold causes the false positive rates for both groups to become similar.
    Panel (b) shows that an increasing degree of fairness corresponds to a steady decrease in accuracy.
    Panel (c) shows the original decision boundary (solid line) and fair decision boundary (dashed line), along with corresponding
    accuracy and false positive rates for groups $z=0$ (crosses) and $z=1$ (circles). Fairness constraints cause the original decision boundary to rotate
    such that previously misclassified examples with $z=0$ are moved into the negative class (decreasing false positives), while well-classified  examples with $z=1$ are
    moved into the positive class (increasing false positives), leading to equal false positive rates for both groups.
    }}
    \vspace{-5mm}
    \label{fig:syn_no_corr}
\end{figure*}

\subsection{Experiments on synthetic data}
In this section, we empirically study the trade-off between fairness and accuracy in a classifier that suffers from  \disc{}.
To this end, we first start with a simple scenario in which the classifier is unfair  in terms of \textit{only} false positive rate \textit{or} false negative rate. Then, we focus on a more complex scenario in which the classifier is unfair in terms of \textit{both}.

\subsubsection{\Disc{} on \textbf{only} false positive rate \textbf{or} false negative rate}\label{sec:syn_disc_single}
\noindent
The first scenario considers a case where a classifier trained on the ground truth data leads to \disc{}  in terms of only the false positive rate (false negative rate), while being fair with respect to false negative rate (false positive rate), \ie, $D_{FPR} \neq 0$ and $D_{FNR} = 0$ (or, alternatively, $D_{FPR} = 0$ and $D_{FNR} \neq 0$).

\vspace{1.5mm}
\xhdr{Experimental setup}
We first generate $10{,}000$ binary class labels ($y \in \{-1, 1\}$) and corresponding sensitive attribute values ($z \in \{0, 1\}$), both uniformly at random, and assign
a two-dimensional user feature vector ($\mathbf{x}$) to each of the points.
To ensure different distributions for negative classes of the two sensitive attribute value groups (so that the two groups have different false positive rates), the user feature vectors are sampled from the following distributions (we sample $2500$ points from each distribution):

\vspace{-4mm}
\begin{align}
p(\mathbf{x}|z=0, y=1) &= \mathcal{N}([2,2], [3, 1; 1, 3]) \nonumber \\
p(\mathbf{x}|z=1, y=1) &= \mathcal{N}([2,2], [3, 1; 1, 3]) \nonumber \\
p(\mathbf{x}|z=0, y=-1) &= \mathcal{N}([1,1], [3, 3; 1, 3]) \nonumber \\
p(\mathbf{x}|z=1, y=-1) &= \mathcal{N}([-2,-2], [3, 1; 1, 3]). \nonumber
\end{align}

\noindent
Next, we train a (unconstrained) logistic regression classifier on this data.
The classifier is able to achieve an accuracy of $0.85$. However, due to difference in feature distributions for the two sensitive attribute value groups, it achieves  $D_{FNR} = 0.14- 0.14= 0$ and $D_{FPR} = 0.25 - 0.06 =0.19$, which constitutes a clear case of \disc{} in terms of false positive rate.

We then train several logistic regression classifiers on the same training data subject to fairness constraints on false positive rate, \ie, we train a logistic regressor by solving problem~\eqref{eq:cons_misclass_convex_logreg}, where $g_{\thetab}(y, \mathbf{x})$ is given by Eq.~\eqref{eq:g_fpr}.
Each classifier constrains the false positive rate covariance ($c$) with a multiplicative factor ($m \in [0,1]$) of the covariance of the unconstrained classifier ($c^*$), that is, $c = mc^*$. Ideally, a smaller $m$, and hence a smaller $c$, would result in more fair outcomes.

\vspace{1.5mm}
\xhdr{Results}
Figure~\ref{fig:syn_no_corr} summarizes the results for this scenario by showing (a) the relation between decision-boundary covariance and the false positive rates for both sensitive attribute values; (b) the trade-off between accuracy and fairness; and (c) the decision boundaries for both the unconstrained classifier (solid) and the fair constrained classifier (dashed).
In this figure, we observe that:
i) as the fairness constraint value $c = mc^*$ goes to zero, the false positive rates for both groups  ($z=0$ and  $z=1$) converge, and hence, the outcomes of the classifier become more fair, \ie, $D_{FPR} \to 0$, while $D_{FNR}$ remains close to zero (the invariance of $D_{FNR}$ may however change depending on the underlying distribution of the data);
ii) ensuring lower values of \disc{} leads to a larger drop in accuracy.

\begin{figure*}[ht]
 \centering

  \subfloat[FPR constraints]{\includegraphics[width=0.6\columnwidth]{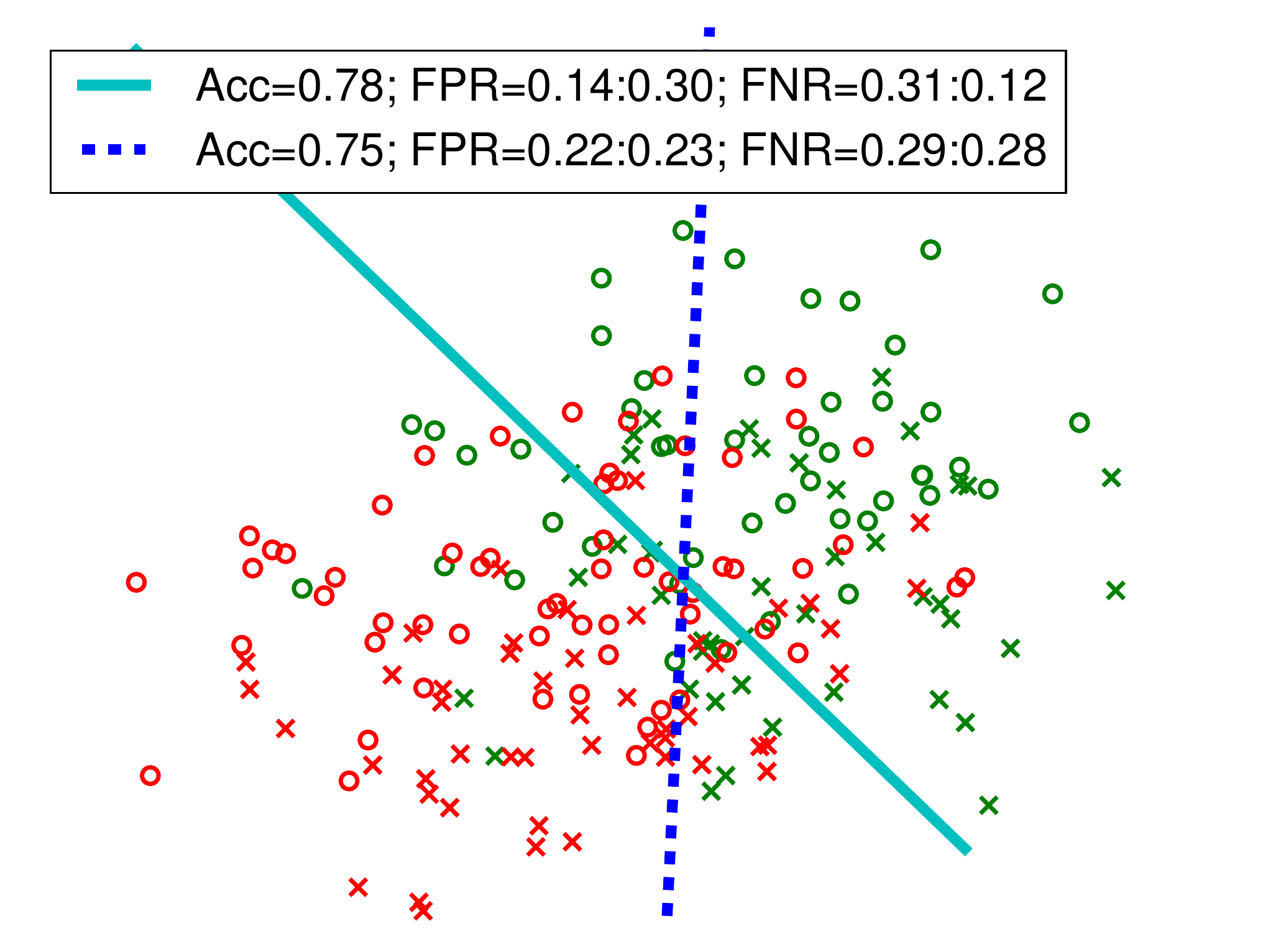}}
  \subfloat[FNR constraints]{\includegraphics[width=0.6\columnwidth]{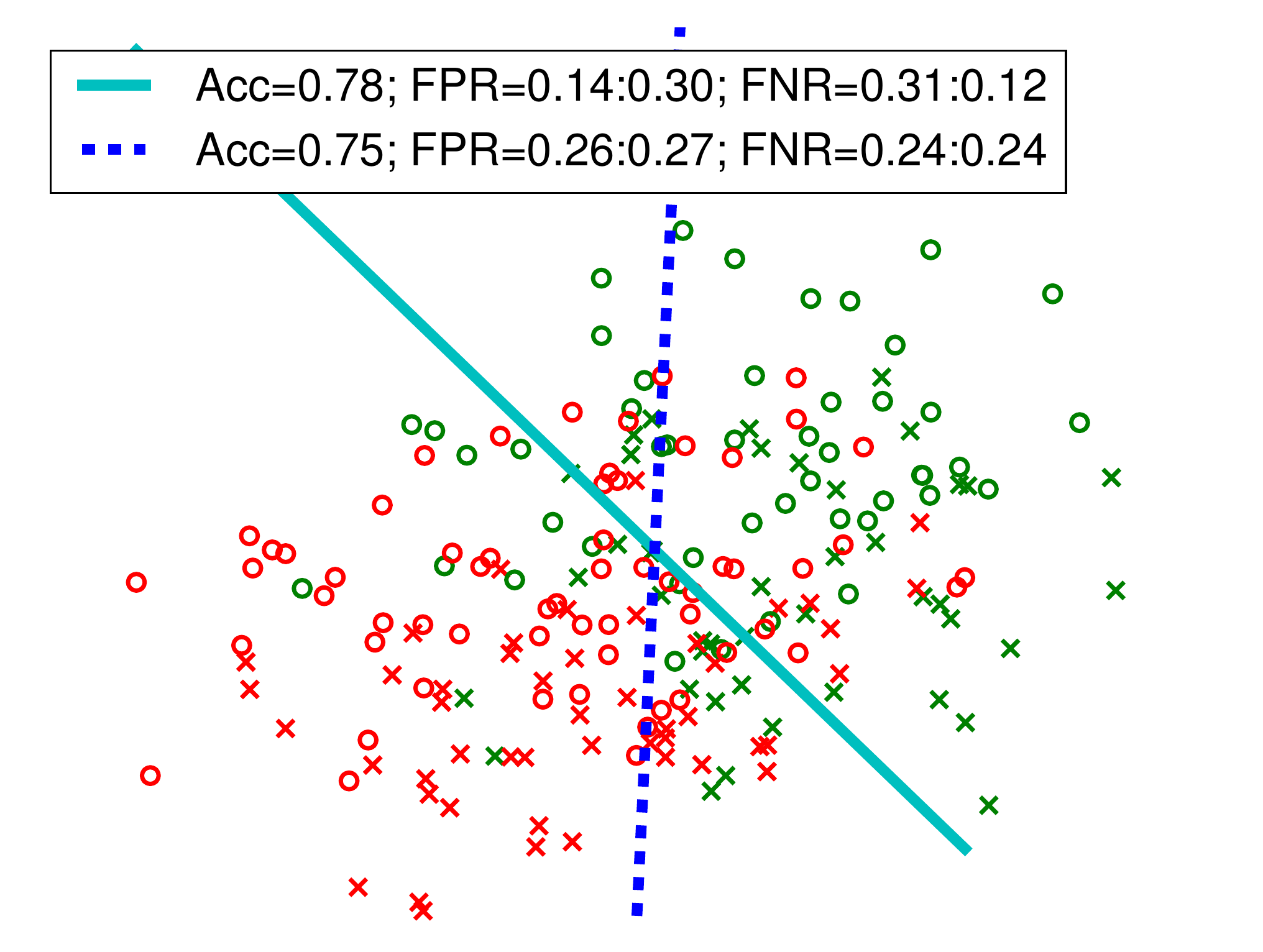}}
  \subfloat[Both constraints]{\includegraphics[width=0.6\columnwidth]{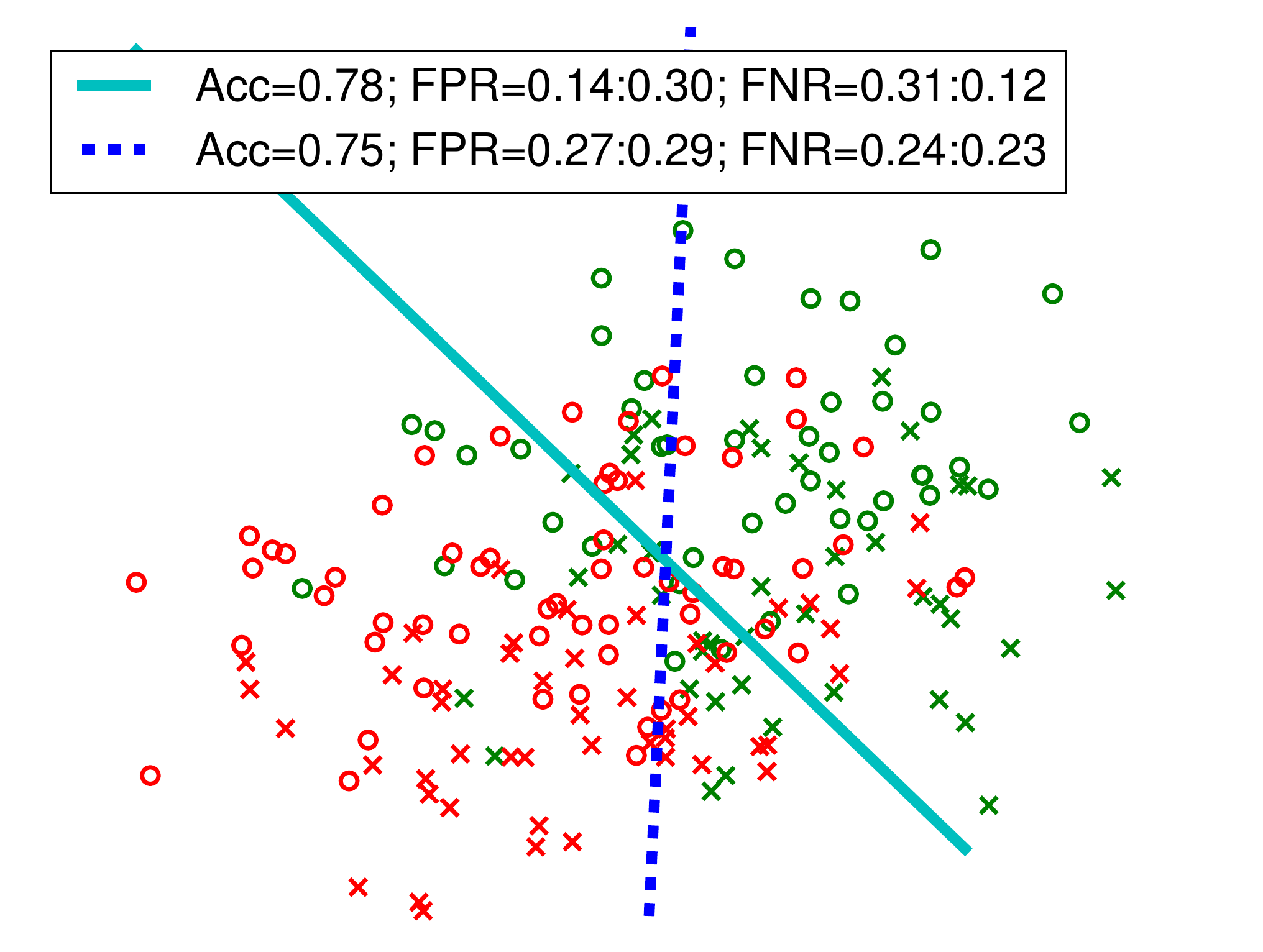}}

  \caption{\small{
    [Synthetic data] $D_{FPR}$ and $D_{FNR}$ have opposite signs. Removing \disc{} on FPR can potentially help remove \disc{} on FNR. Removing \disc{} on both at the same time leads to very similar results.
  }}
  \label{fig:syn_opposite_fp_fn}
    \vspace{-4mm}
\end{figure*}

\begin{figure*}[t]
 \centering

  \subfloat[FPR constraints]{\includegraphics[width=0.6\columnwidth]{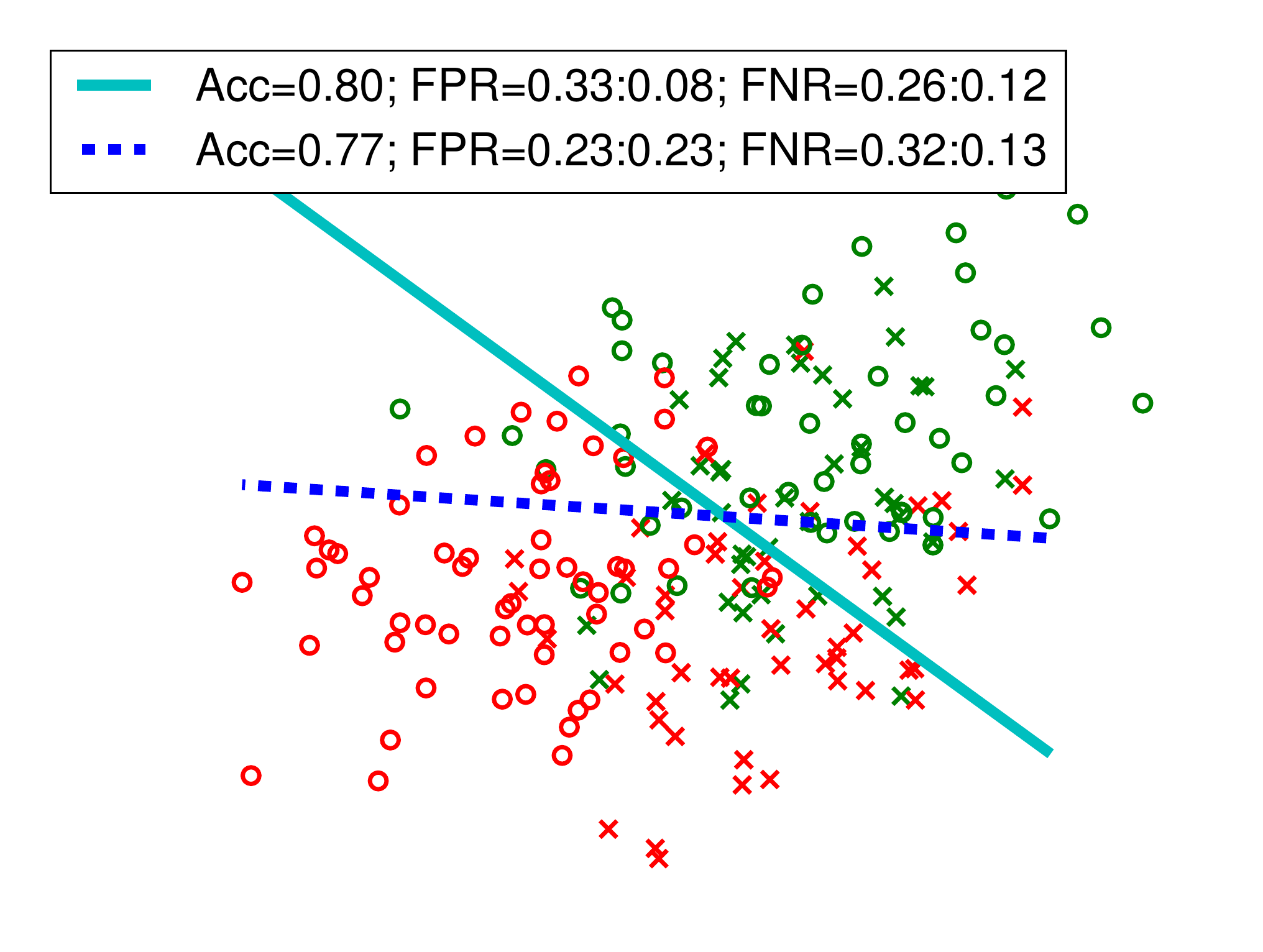}}
  \label{fig:syn_same_fp_fn_fpr}
  \subfloat[FNR constraints]{\includegraphics[width=0.6\columnwidth]{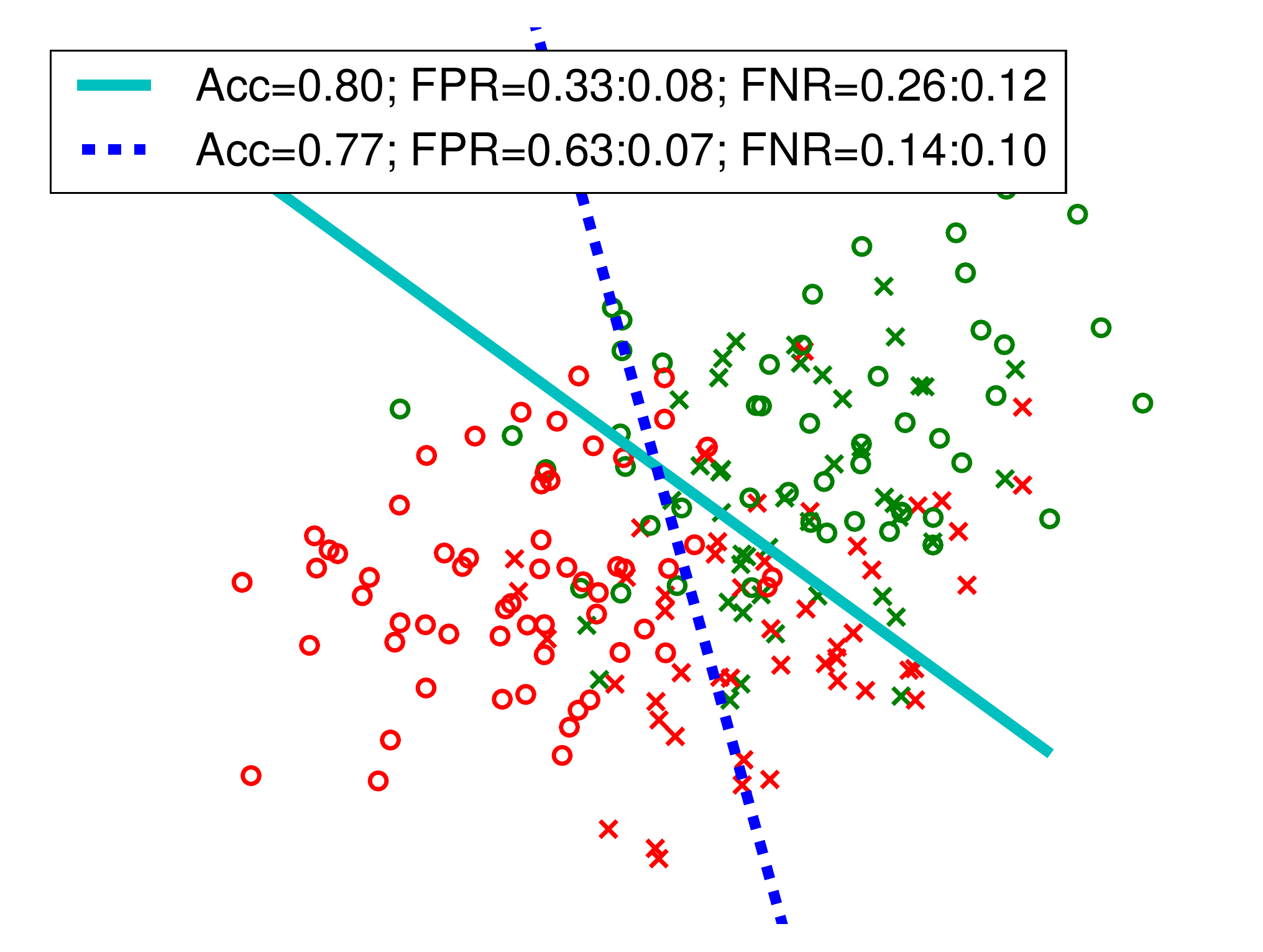}}
  \label{fig:syn_same_fp_fn_fnr}
  \subfloat[Both constraints]{\includegraphics[width=0.6\columnwidth]{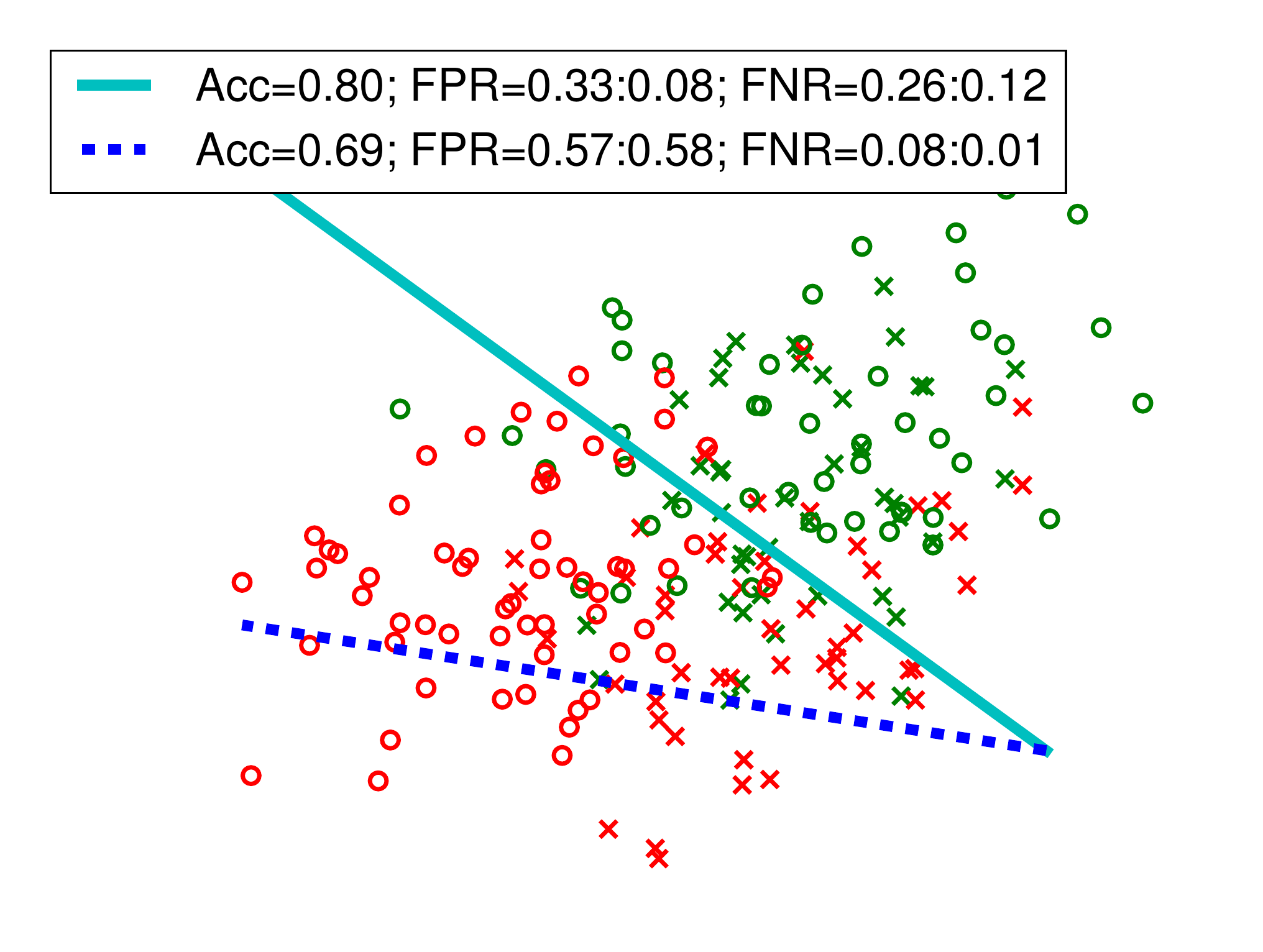}}
  \label{fig:syn_same_fp_fn_both}

  \caption{\small{
    [Synthetic data] $D_{FPR}$ and $D_{FNR}$ have the same sign. Removing \disc{} on FPR can potentially increase \disc{} on FNR. Removing \disc{} on both at the same time  causes a larger drop in accuracy.
  }}
  \label{fig:syn_same_fp_fn}
    \vspace{-4mm}
\end{figure*}

\subsubsection{\Disc{} on \textbf{both} false positive rate and false negative rate}\label{sec:syn_both_disc}
\noindent
In this section, we consider a more complex scenario,
where the outcomes of the classifier suffer from \disc{}  with respect to \textit{both} false positive rate and false negative rate, \ie, both $D_{FPR}$ and $D_{FNR}$ are non-zero.
This scenario can in turn be split into two cases:

\xhdr{I}
$D_{FPR}$ and $D_{FNR}$ have \textit{opposite signs}, \ie, the decision boundary disproportionately \textit{favors} subjects from a certain sensitive attribute value group to be in the positive class (even when such assignments are misclassifications) while disproportionately assigning the subjects from the other group to the negative class. As a result, false positive rate for one group is higher than the other, while the false negative rate for the same group is lower.
\\
\xhdr{II}
$D_{FPR}$ and $D_{FNR}$ have the \textit{same sign}, \ie, both false positive as well as false negative rate are higher for a certain sensitive attribute value group. These cases might arise in scenarios when a certain group is harder to classify than the other.

\noindent
Next, we experiment with each of the above cases separately.

\vspace{1.5mm}
\noindent \emph{--- \bf Case I:}
To simulate this scenario, we first generate $2{,}500$ samples from each of the  following distributions:
\begin{align}
p(\mathbf{x}|z=0, y=1) &= \mathcal{N}([2,0], [5, 1; 1, 5]) \nonumber \\
p(\mathbf{x}|z=1, y=1) &= \mathcal{N}([2,3], [5, 1; 1, 5]) \nonumber \\
p(\mathbf{x}|z=0, y=-1) &= \mathcal{N}([-1,-3], [5, 1; 1, 5]) \nonumber \\
p(\mathbf{x}|z=1, y=-1) &= \mathcal{N}([-1,0], [5, 1; 1, 5]) \nonumber
\end{align}
An unconstrained logistic regression classifier on this dataset attains an overall accuracy of $0.78$ but leads to a false positive rate of $0.14$ and $0.30$ (\ie, $D_{FPR} = 0.14 - 0.30 =-0.16$) for the sensitive attribute groups $z=0$ and $z=1$, respectively; and  false negative rates of $0.31$ and $0.12$ (\ie, $D_{FNR} = 0.31 - 0.12 = 0.19$).
Finally, we train three different fair classifiers, with fairness constraints on (i) false positive rates---$g_{\thetab}(y, \mathbf{x})$  given by Eq.~\eqref{eq:g_fpr}, (ii) false negative rates---$g_{\thetab}(y, \mathbf{x})$  given by Eq.~\eqref{eq:g_fnr} and (iii) on both false positive and false negative rates---separate constraints for $g_{\thetab}(y, \mathbf{x})$ given by Eq.~\eqref{eq:g_fpr} and Eq.~\eqref{eq:g_fnr}.

\vspace{1.5mm}
\xhdr{Results}
Figure~\ref{fig:syn_opposite_fp_fn} summarizes the results for this scenario by showing the decision boundaries for the unconstrained classifier (solid) and the constrained fair classifiers.
Here, we can observe several interesting patterns.
First, removing \disc{} on only false positive rate causes a rotation in the decision boundary to move previously  \emph{misclassified}  examples with $z=1$ into the negative class, \emph{decreasing} their false positive rate. However, in the process, it also moves previously \emph{well-classified} examples with $z=1$ into the negative class, \emph{increasing} their false negative rate.
As a consequence, controlling \disc{} on false positive rate (Figure~\ref{fig:syn_opposite_fp_fn}(a)), also  removes \disc{} on false negative rate.
A similar effect occurs when we control \disc{} only with respect to the false negative rate (Figure~\ref{fig:syn_opposite_fp_fn}(b)), and therefore, provides similar results as the constrained classifier for both false positive and false negative rates (Figure~\ref{fig:syn_opposite_fp_fn}(c)).
This effect is explained by the distribution of the data, where the centroids of the clusters for the group with $z=0$ are shifted with respect to the ones for the group $z=1$.

\vspace{1.5mm}
\noindent \emph{--- \bf Case II:}
To simulate the scenario where both $D_{FPR}$ and $D_{FNR}$ have the same sign, we generate $2{,}500$ samples from each
of the following distributions:
\begin{align}
p(\mathbf{x}|z=0, y=1) &= \mathcal{N}([1,2], [5, 2; 2, 5]) \nonumber \\
p(\mathbf{x}|z=1, y=1) &= \mathcal{N}([2,3], [10, 1; 1, 4]) \nonumber \\
p(\mathbf{x}|z=0, y=-1) &= \mathcal{N}([0,-1], [7, 1; 1, 7]) \nonumber \\
p(\mathbf{x}|z=1, y=-1) &= \mathcal{N}([-5,0], [5, 1; 1, 5]) \nonumber
\end{align}
Then, we train an unconstrained logistic regression classifier on this dataset. It attains an accuracy of $0.80$ but leads to $D_{FPR} = 0.33- 0.08 = 0.25$ and $D_{FNR} = 0.26-0.12 = 0.14$, resulting in  \disc{} in terms of both false positive and negative rates.
Then, similarly to the previous scenario, we train three different kind of constrained classifiers to remove \disc{} on (i) false positive rate, (ii) false negatives rate, and (iii) both.

\vspace{1.5mm}
\xhdr{Results} Figure~\ref{fig:syn_same_fp_fn} summarizes the results by showing  the decision boundaries for both the unconstrained classifiers (solid) and the fair constrained classifier (dashed) when controlling for \disc{} with respect to false positive rate, false negative rate and both, respectively.
We  observe several interesting patterns. First, controlling \disc{} for only false positive rate (false negative rate), leads to a minor drop in accuracy, but can exacerbate the \disc{} on false negative rate (false positive rate).
For example, while the decision boundary is moved to control for \disc{} on false negative rate, that is, to ensure that more examples with $z=0$ are well-classified in the positive class (reducing false negative rate), it also moves previously well-classified negative examples into the positive class, hence increasing the false positive rate.
A similar phenomenon occur when controlling \disc{} with respect to only false positive rate.
As a consequence, controlling for both types of \disc{} simultaneously brings $D_{FPR}$ and $D_{FNR}$ close to zero, but causes a large drop in accuracy.

\subsubsection{Performance Comparison}
\noindent
In this section, we compare the performance of our scheme with two different methods on the synthetic datasets described above.
In particular, we compare the performance of the following approaches:

\noindent
\textbf{\textit{Our method}:} implements our scheme to avoid disparate treatment and
disparate mistreatment \textit{simultaneously}. \Disc{} is avoided by using fairness constraints
(as described in Sections~\ref{sec:syn_disc_single} and~\ref{sec:syn_both_disc}). Disparate treatment
is avoided by ensuring that sensitive attribute information is not used while making decisions, \ie, by
keeping user feature vectors ($\mathbf{x}$) and
the sensitive features ($z$) disjoint.
All the explanatory simulations on synthetic data shown earlier (Sections~\ref{sec:syn_disc_single} and~\ref{sec:syn_both_disc}) implement this scheme.

\noindent
\textbf{\textit{Our method\textsubscript{sen}}:}  implements our scheme to avoid disparate
mistreatment only. The user feature vectors ($\mathbf{x}$) and the sensitive features ($z$) are not
disjoint, that is, $z$ is used as a learnable feature. Therefore, the sensitive attribute information is
used for decision making, resulting in disparate treatment.

\noindent
\textbf{\textit{Hardt et al.}}~\cite{hardt_nips16}: operates by post-processing the outcomes of
an unfair classifier (logistic regression in this case) and using different decision thresholds for different sensitive attribute value groups
to achieve fairness. By construction, it needs the sensitive attribute information while making decisions,
and hence cannot avoid disparate treatment.

\noindent
\textbf{\textit{Baseline}:} tries to remove \disc{} by introducing different penalties for misclassified
data points with different sensitive attribute values during training phase. Specifically, it proceeds in two steps.
First, it trains an (unfair) classifier minimizing a loss function (\eg, logistic loss)
 over the training data. Next, it selects the set of misclassified data points from the sensitive attribute value group
 that presents the higher error rate. For example, if one wants to remove \disc{} with respect to false positive
rate and $D_{FPR}>0$ (which means the false positive rate for points with $z=0$ is higher than that of $z=1$), it
selects the set of misclassified data points in the training set having $z = 0$ and  $y = -1$. Next, it iteratively re-trains
the classifier with increasingly higher penalties on this set of data points until a certain fairness level is achieved
in the training set (until $D_{FPR}\leq \epsilon$).
The algorithm is summarized in Figure~\ref{alg:baseline}, particularized to ensure fairness in terms of false positive
rate. This process can be intuitively extended to account for fairness in terms of false negative rate or for \textit{both} false
positive rate and false negative rate.
Like \textbf{\textit{Our method}}, the baseline does not use sensitive attribute information while making decisions.

\RestyleAlgo{}
\DontPrintSemicolon
\begin{algorithm}[t]
\SetKwInput{Initialize}{Initialize}
\KwInput{Training set $\mathcal{D}=\{ (\mathbf{x}_i, y_i, z_i) \}_{i=1}^{N}$, $\Delta>0$ $\epsilon>0$}
\KwOutput{Fair baseline decision boundary $\thetab$}
 \Initialize{Penalty $C= 1$}

Train (unfair) classifier $\thetab ={\argmin}_{\thetab} \sum_{\mathbf{d} \in \mathcal{D}} L(\thetab, \mathbf{d})$

Compute $\hat{y}_i= \mathrm{sign} (d_{\thetab} (\mathbf{x}_i))$ and $D_{FP}$ on $\mathcal{D}$.

\lIf{$D_{FP}>0$}{
$s=0$
} \lElse{
$s=1$
}

$\mathcal{P}= \{ \mathbf{x}_i, y_i, z_i | \hat{y}\neq y_i, z_i = s\}$, $\bar{\mathcal{P}} = \mathcal{D} \setminus \mathcal{P}$.

 \While{ $D_{FP}>\epsilon$}{

  Increase penalty: $C=C+\Delta$.

  $\thetab =  {\argmin}_{\thetab}  C \sum_{\mathbf{d} \in \mathcal{P}} L(\thetab, \mathbf{d}) +  \sum_{\mathbf{d} \in \bar{\mathcal{P}}} L(\thetab, \mathbf{d})$

   }
\caption{Baseline method for removing disparate  mistreatment on false positive rates.} \label{alg:baseline}
\vspace{-6ex}
\end{algorithm}

\xhdr{Comparison results}
Table~\ref{table:comparison_results} shows the performance comparison for all the methods on the three synthetic datasets described above.
We can observe that, while all four methods mostly achieve similar levels of fairness, they do it at different costs in terms of accuracy.
Both \textbf{\textit{Our method\textsubscript{sen}}} and \textbf{\textit{Hardt et al.}}---which use sensitive feature information while
making decisions---present the best performance in terms of accuracy (due to the additional information available to them).
However, as explained earlier, these two methods suffer from disparate treatment.
On the other hand, the implementation of our scheme to simultaneously remove disparate mistreatment and disparate treatment (\textbf{\textit{Our method}}) does so with further accuracy drop of only $\sim$$5\%$ with respect to the above two methods that cause
 disparate treatment.
Finally, the \textbf{\textit{baseline}} is sometimes unable to achieve fairness. When it does achieves fairness, it does so
at a (sometimes much) greater cost in accuracy in comparison with the competing methods.

In summary, our method achieves the same performance as \textit{Hardt et al.} when making use of the same information in the
data, \ie, non-sensitive as well as sensitive features.  However, in contrast to \textit{Hardt et al.}, it also allows us to simultaneously remove both disparate mistreatment and disparate treatment at a small additional cost in terms of accuracy.

\begin{table*}[ht]
\centering
\resizebox{\textwidth}{!}{
\def\arraystretch{1.2}

\begin{tabular}{c c  c | c | c | c || c | c | c || c | c | c |}

\cline{4-12}
&&&
\multicolumn{3}{ c|| }{\textbf{FPR constraints}}&
\multicolumn{3}{ c|| }{\textbf{FNR constraints}}&
\multicolumn{3}{ c| }{\textbf{Both constraints}}
\\ \cline{4-12}
&&&
 $\mathbf{Acc.}$ &  $\mathbf{D_{FPR}}$ &  $\mathbf{D_{FNR}}$ &  $\mathbf{Acc.}$ &  $\mathbf{D_{FPR}}$ &  $\mathbf{D_{FNR}}$ &  $\mathbf{Acc.}$ &  $\mathbf{D_{FPR}}$ &  $\mathbf{D_{FNR}}$
\\  \cline{4-12}
\vspace*{-0.5ex}
\\ \cline{1-2} \cline{4-12}
\multicolumn{1}{ |c|  }{   \multirow{4}{*}{\shortstack{\textbf{Synthetic}\\\textbf{setting 1} \\ (Figure~\ref{fig:syn_no_corr}) }}   } &
\multicolumn{1}{ c|  }{  \textbf{Our method}}  &&
$0.80$ 	 & 	 $0.02$ 	 & 	 $0.00$ 	 &
$-$      &   $-$         &   $-$         &
$-$      &   $-$         &   $-$
\\ \cline{2-2} \cline{4-12}
\multicolumn{1}{ |c|  }{} &
\multicolumn{1}{ c|  }{  \textbf{Our method\textsubscript{sen}}}  &&
$0.85$ 	 & 	 $0.00$ 	 & 	 $0.25$ 	 &
$-$      &   $-$         &   $-$         &
$0.83$ 	 & 	 $0.07$ 	 & 	 $0.01$
\\ \cline{2-2} \cline{4-12}
\multicolumn{1}{ |c|  }{} &
\multicolumn{1}{ c|  }{  \textbf{Baseline}}  &&
$0.65$ 	 & 	 $0.00$ 	 & 	 $0.00$ 	 &
$-$      &   $-$         &   $-$         &
$-$      &   $-$         &   $-$
\\ \cline{2-2} \cline{4-12}
\multicolumn{1}{ |c|  }{} &
\multicolumn{1}{ c|  }{  \textbf{Hardt et al.}}  &&
$0.85$ 	 & 	 $0.00$ 	 & 	 $0.21$ 	 &
$-$      &   $-$         &   $-$         &
$0.80$ 	 & 	 $0.00$ 	 & 	 $0.02$
\\ \cline{1-2} \cline{4-12}
\vspace*{-0.5ex}
\\ \cline{1-2} \cline{4-12}
\multicolumn{1}{ |c|  }{   \multirow{4}{*}{\shortstack{\textbf{Synthetic}\\ \textbf{setting 2} \\ (Figure~\ref{fig:syn_opposite_fp_fn}) }}   } &
\multicolumn{1}{ c|  }{  \textbf{Our method}}  &&
$0.75$   &   $-0.01$   &   $0.01$    &
$0.75$   &   $-0.01$   &   $0.01$    &
$0.75$   &   $-0.01$   &   $0.01$
\\ \cline{2-2} \cline{4-12}
\multicolumn{1}{ |c|  }{} &
\multicolumn{1}{ c|  }{  \textbf{Our method\textsubscript{sen}}}  &&
$0.80$   &   $0.00$    &   $0.03$    &
$0.80$   &   $0.02$    &   $0.01$    &
$0.80$   &   $0.01$    &   $0.02$
\\ \cline{2-2} \cline{4-12}
\multicolumn{1}{ |c|  }{} &
\multicolumn{1}{ c|  }{  \textbf{Baseline}}  &&
$0.59$   &   $-0.01$   &   $0.15$    &
$0.59$   &   $-0.15$   &   $0.01$    &
$0.76$   &   $-0.04$   &   $0.03$
\\ \cline{2-2} \cline{4-12}
\multicolumn{1}{ |c|  }{} &
\multicolumn{1}{ c|  }{  \textbf{Hardt et al.}}  &&
$0.80$   &   $0.00$   &   $0.03$     &
$0.80$   &   $0.03$    &   $0.00$    &
$0.79$   &   $0.00$    &   $-0.01$
\\ \cline{1-2} \cline{4-12}
\vspace*{-0.5ex}
\\ \cline{1-2} \cline{4-12}
\multicolumn{1}{ |c|  }{   \multirow{4}{*}{\shortstack{\textbf{Synthetic} \\ \textbf{setting 3} \\ (Figure~\ref{fig:syn_same_fp_fn}) }}   } &
\multicolumn{1}{ c|  }{  \textbf{Our method}}  &&
$0.77$   &   $0.00$    &   $0.19$    &
$0.77$   &   $0.55$    &   $0.04$    &
$0.69$   &   $-0.01$   &   $0.06$
\\ \cline{2-2} \cline{4-12}
\multicolumn{1}{ |c|  }{} &
\multicolumn{1}{ c|  }{  \textbf{Our method\textsubscript{sen}}}  &&
$0.78$   &   $0.00$    &   $0.42$    &
$0.79$   &   $0.38$    &   $0.03$    &
$0.77$   &   $0.14$    &   $0.06$
\\ \cline{2-2} \cline{4-12}
\multicolumn{1}{ |c|  }{} &
\multicolumn{1}{ c|  }{  \textbf{Baseline}}  &&
$0.57$   &   $0.01$    &   $0.09$    &
$0.67$   &   $0.44$    &   $0.01$    &
$0.38$   &   $-0.43$   &   $0.01$
\\ \cline{2-2} \cline{4-12}
\multicolumn{1}{ |c|  }{} &
\multicolumn{1}{ c|  }{  \textbf{Hardt et al.}}  &&
$0.78$   &   $0.01$    &   $0.44$    &
$0.79$   &   $0.41$    &   $0.02$    &
$0.67$   &   $0.02$    &   $0.00$
\\ \cline{1-2} \cline{4-12}
\vspace*{-0.5ex}
\\ \cline{1-2} \cline{4-12}
\multicolumn{1}{ |c|  }{   \multirow{3}{*}{\shortstack{\textbf{ProPuclica} \\ \textbf{COMPAS} \\ (Section~\ref{sec:propublica}) }}   } &
\multicolumn{1}{ c|  }{  \textbf{Our method\textsubscript{sen}}}  &&
$0.660$    &   $0.06$    &   $-0.14$  &
$0.662$    &   $0.03$    &   $-0.10$  &
$0.661$    &   $0.03$    &   $-0.11$
\\ \cline{2-2} \cline{4-12}
\multicolumn{1}{ |c|  }{} &
\multicolumn{1}{ c|  }{  \textbf{Baseline}}  &&
$0.643$    &   $0.03$    &   $-0.11$  &
$0.660$    &   $0.00$    &   $-0.07$  &
$0.660$    &   $0.01$    &   $-0.09$
\\ \cline{2-2} \cline{4-12}
\multicolumn{1}{ |c|  }{} &
\multicolumn{1}{ c|  }{  \textbf{Hardt et al.}}  &&
$0.659$    &   $0.02$    &   $-0.08$  &
$0.653$    &   $-0.06$   &   $-0.01$  &
$0.645$    &   $-0.01$   &   $-0.01$
\\ \cline{1-2} \cline{4-12}
\end{tabular}
}

\caption{
Performance of different methods while removing \disc{} with respect to false positive rate, false negative rate and both.
}
\vspace{3mm}
\label{table:comparison_results}
\end{table*}
\subsection{Real world dataset: ProPublica COMPAS} \label{sec:propublica}
\noindent
In this section, we experiment with the COMPAS risk assessment dataset compiled by ProPublica~\cite{propublica_compas}
and show that our method can significantly reduce disparate mistreatment at a modest cost in terms of accuracy.

\vspace{1.5mm}
\xhdr{Dataset and experimental setup}
ProPublica compiled a list of all criminal offenders screened through the COMPAS (Correctional Offender Management Profiling for Alternative Sanctions) tool
\footnote{
\scriptsize{COMPAS tries to predict the recidivism risk (on a scale of 1--10) of a criminal offender by analyzing answers to 137 questions pertaining to the offender's criminal history and behavioral patterns~\cite{compas_questions}.}
}
in Broward County, Florida during 2013-2014. The data includes information on the offenders' demographic features (gender, race,
age), criminal history (charge for which the person was arrested, number of prior offenses) and the risk score assigned to the offender
by COMPAS. ProPublica also collected the \emph{ground truth} on whether or not these individuals actually recidivated within two
years after the screening.
For more information about the data collection, we point the reader to a detailed description~\cite{propublica_methodology}.
Some of the follow-up discussion on this dataset can be found at~\cite{propublica_response,flores_rejoinder}.

In this analysis, for simplicity, we only consider a subset of offenders whose race was either black or white. Recidivism rates for the two groups are shown in Table~\ref{table:compas-stats}.

Using this ground truth, we build an unconstrained logistic regression classifier to predict whether an offender will (positive class) or
will not (negative class) recidivate within two years. The set of features used in the classification task are described
in Table~\ref{table:compas-features}.
\footnote{
\scriptsize{Notice that goal of this section is not to analyze the best set of features for recidivism prediction, rather, we
focus on showing that our method can effectively remove \disc{} in a given dataset. Hence, we chose to use the same
set of features as used by ProPublica for their analysis.}
}${}^{,}$
\footnote{
\scriptsize{Since race is one of the features in the learnable set, we additionally assume that \textit{all} the methods have
access to the sensitive attributes while making decisions.}
}

The (unconstrained) logistic regression classifier leads to an accuracy of $0.668$.
However, the classifier yields  false positive rates of $0.35$ and $0.17$, respectively, for  blacks and whites (\ie, $D_{FPR} = 0.18$), and false negative rates of $0.31$ and $0.61$ (\ie, $D_{FNR} = -0.30$).
These results constitute a clear case of \disc{} in terms of both false positive rate and false negative rate.
The classifier puts one  group (blacks) at relative disadvantage  by disproportionately misclassifying negative (did not recidivate) examples from this group into positive (did recidivate) class. This disproportional assignment results in a significantly higher false positive rate for blacks as compared to whites. On the other hand, the classifier puts the other group (whites) on a relative advantage by disproportionately misclassifying positive (did recidivate) examples from this group into negative (did not recidivate) class (resulting in a higher false negative rate).
Note that this scenario resembles our synthetic example Case I in Section \ref{sec:syn_both_disc}.

Finally, we train logistic regression classifiers with three types of constraints: constraints on false positive rate, false negative rate, and on both.

\vspace{1.5mm}
\xhdr{Results}
Table~\ref{table:comparison_results} (last block) summarizes the results by showing the trade-off between fairness and accuracy achieved by our method, the method by Hardt et al., and the baseline.
Similarly to the results in Section \ref{sec:syn_both_disc}, we observe that for all three mehtods, controlling for \disc{} on false positive rate (false negative rate) also helps decrease \disc{} on false negative rate (false positive rate). Moreover, all three methods are able to
achieve similar accuracy for a given level of fairness.

Additionally, we observe that our method (as well as the baseline) does not completely remove \disc{}, \ie, it does not achieve zero $D_{FPR} $ or/and $D_{FNR} $ in any of the cases.
This is probably due to the relatively small size of the dataset
\footnote{
\scriptsize{$2,639$ examples in the training set.}
}
(and hence a smaller ratio between number of training examples and number of learnable features), which hinders a
robust estimate of misclassification covariance (Eqs. \ref{eq:g_fpr} and \ref{eq:g_fnr}). This highlights the fact that our
method can suffer from reduced performance on small datasets.
In scenarios with sufficiently large training datasets, we expect more reliable estimates of covariance, and hence, a better
performance from our method.
On the other hand, the method by Hardt et al. is able to achieve both zero $D_{FPR} $ and $D_{FNR} $ while controlling for
\disc{} on both false positive and false negative rates (Table~\ref{table:comparison_results})---albeit at a considerable drop
in terms of accuracy. Since this method operates on a data of much smaller dimensionality (the final classifier probability estimates),
it is not expected to suffer as much from the small size of the dataset as compared to our method or the baseline (which depend
on the misclassification covariance computed on the feature set).

\begin{table}
\centering
\begin{tabular}{c|c|c|c}
\hline
Race  &  Yes        & No        & Total \\ \hline
Black &  $ 1,661 (52\%)  $ & $ 1,514 (48\%)$ &  $ 3,175 (100\%)$ \\
White &  $ 8,22  (39\%) $  & $1,281  (61\%)$ &  $ 2,103 (100\%)$ \\ \hline
Total &  $ 2,483 (47\%) $  & $2,795  (53\%)$ &  $ 5,278 (100\%)$ \\\hline
\end{tabular}
\caption{Recidivism rates in ProPublica COMPAS data for both races.} \label{table:compas-stats}
\vspace{3mm}
\end{table}

\begin{table}
\centering
\begin{tabular}{c|c}
\hline
Feature  &  Description \\ \hline
Age Category &  $<25$, between $25$ and $45$, $>45$ \\
Gender &  Male or Female \\
Race &  White or Black \\
Priors Count &  0--37 \\
Charge Degree &  Misconduct or Felony \\
\hline
2-year-rec. & Whether (+ve) or not (-ve) the  \\
(target feature)  & defendant recidivated within two years
\end{tabular}
\caption{Description of features used from ProPublica COMPAS data.} \label{table:compas-features}
\end{table}

\section{Discussion and future work}
\noindent
As shown in Section~\ref{sec:eval}, the method proposed in this paper provides a flexible tradeoff between disparate mistreatment-based fairness and accuracy.
It also allows to avoid \disc{} and disparate treatment \textit{simultaneously}. This feature might be specially
useful in scenarios when the sensitive attribute information is not available (\eg, due to privacy reasons) or is prohibited from being
used due to disparate treatment laws~\cite{barocas_2016}.

Although we proposed fair classifier formulations to remove \disc{} only on false positive and false negative rates, as described in Section~\ref{sec:format_setting}, \disc{} can also be measured with respect to false discovery and false omission rates.
Extending our current formulation to include false discovery and false omission rates is a non-trivial task due to computational complexities involved.
A natural extension of this work would be to include these other measures of \disc{} into our fair classifier formulation.

Finally, we
would like to point out that the current formulation of fairness constraints may suffer from the following limitations.
Firstly, the proposed formulation to train fair classifiers is not a convex program, but a disciplined convex-concave program (DCCP), which can be efficiently solved using heuristic-based methods~\cite{boyd_concave_convex}.  While these methods are shown
to work well in practice, unlike convex optimization, they do not provide any guarantees on the global optimality of the solution.
Secondly, since com\-pu\-ting the analytical covariance in fairness constraints is not a trivial task, we approximate it through Monte Carlo covariance on the training set (Eq.~\ref{eq:fairness-proxy}). While this approximation is expected to work well when a reasonable amount of training data is provided, it might be inaccurate for smaller datasets.

\bibliographystyle{abbrv}

\end{document}